 \documentclass[final,5p,times,twocolumn,authoryear]{melsarticle}

\usepackage{amssymb}

\usepackage{amsmath}
\usepackage{lipsum}
\usepackage{wasysym}
\usepackage{booktabs}
\usepackage{multirow}
\usepackage[section]{placeins}

\begin{document}

\begin{frontmatter}



\title{Dynamic Domain Discrepancy Adjustment for Active Multi-Domain Adaptation}



\corref{cor1}
\cortext[cor1]{Corresponding author}


\author[add1] {Long Liu}
\ead{liulong@xaut.edu.cn}
\author[add1]{Bo Zhou} 
\ead{1220311012@stu.xaut.edu.cn}
\author[add1]{Zhipeng Zhao}
\ead{2220320119@stu.xaut.edu.cn}
\author[add1]{Zening Liu}   
\ead{znliu@stu.xaut.edu.cn}
\address[add1]{Xi'an University of Technology, Xi'an , 710048 , China}

\begin{abstract}
Multi-source unsupervised domain adaptation (MUDA) aims to transfer knowledge from related source domains to an unlabeled target domain. While recent MUDA methods have shown promising results, most of them focus on aligning the overall feature distributions across source domains, which can lead to negative effects due to redundant features within each domain. Moreover, there is a significant performance gap between MUDA and supervised methods. To address these challenges, we propose a novel approach called Dynamic Domain Discrepancy Adjustment for Active Multi-Domain Adaptation (D$^3$AAMDA). Firstly, we establish a multi-source dynamic modulation mechanism during the training process based on the degree of distribution differences between source and target domains. This mechanism controls the alignment level of features between each source domain and the target domain and effectively use the local advantageous feature information within the source domains. Additionally, we propose a Multi-source Active Boundary Sample Selection (MABS) strategy, which utilizes a guided dynamic boundary loss to design an efficient query function for selecting important samples. This strategy achieves improved generalization to the target domain with minimal sampling costs. We extensively evaluate our proposed method on commonly used domain adaptation datasets, comparing it against existing UDA and ADA methods. The experimental results unequivocally demonstrate the superiority of our approach.
\end{abstract}




\end{frontmatter}




\section{Introduction}
\label{introduction}

Deep neural networks have demonstrated exceptional performance when trained on large annotated datasets. However, obtaining the desired annotated data in the real world is expensive, and models trained on such datasets may struggle to generalize to new datasets and tasks due to domain shift. Unsupervised Domain Adaptation (UDA) techniques have been developed to mitigate the impact of domain shift on model performance when only unlabeled data is available in the target domain.

Early UDA methods primarily focused on single-source scenarios. However, in practical applications, multiple source domains relevant to the target domain are often accessible. Leveraging multiple source domains to enhance UDA performance is desirable. Nevertheless, due to different data collection conditions in each source domain, feature distribution discrepancies exist between each source domain and the target domain. Simply merging multiple source domain data and performing single-source domain adaptation may exacerbate domain shift. Therefore, the key to addressing multi-source domain adaptation lies in establishing an optimal domain adaptation learning process that considers the diverse feature distribution differences between each source domain and the target domain, as well as the approaches for domain transfer.

To address the aforementioned issues , some Multi-source Unsupervised Domain Adaptation (MUDA) methods \citep{a13,a14,a16,a37} adopt a holistic approach to align multiple source domains with the target domain and employ weighted classification strategies at the decision level. However, these methods aim to map features from the source and target domains to a shared feature space, which may result in the loss of domain-specific representations and negative transfer. Another category of methods \citep{a43,a45} focuses on aligning inter-class features by aligning the classes of the source and target domains to obtain an optimal classifier for target domain samples. However, due to distribution differences, achieving complete alignment of class boundaries between each source domain and the target domain is challenging.

Although the methods above have demonstrated remarkable performance, they still lack sufficient consideration of the distribution differences among source domains. As depicted in Figure 1, the distribution of multi-source data is complex, with varying degrees of correlation between the overall distribution of source domains and the target domain, as well as local distribution correlations.

\begin{figure}
	\centering 
	\includegraphics[width=0.4\textwidth, angle=360]{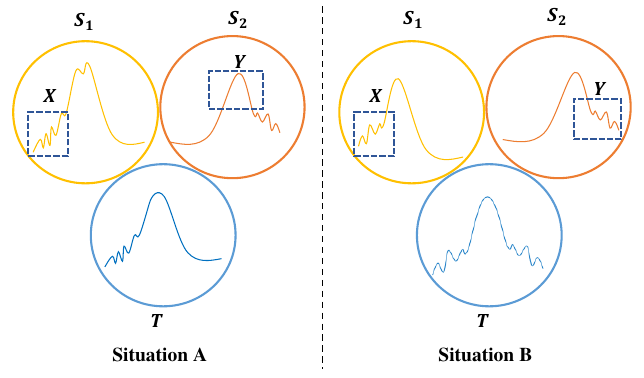}	
    \setlength{\abovecaptionskip}{2pt}%
	\caption{The local distribution relationships between source and target domains: in situation A, source domain $S_1$ is closer to the target domain in the overall distribution, while source domain $S_1$ exhibits similarity in the X segment and source domain $S_2$ exhibits similarity in the Y segment in terms of local distribution patterns. In situation B, both source domains $S_1$ and $S_2$ have similar differences in data distribution compared to the target domain, with source domain $S_1$ showing similarity in the X segment and source domain $S_2$ showing similarity in the Y segment.} 
\end{figure}

In more practical application scenarios, although acquiring complete labels for the target domain is challenging, it is possible to improve domain adaptation (DA) performance by selecting a small portion of target domain data for annotation within a given budget. Active learning (AL) techniques aim to select representative samples from a large pool of unlabeled samples for annotation and model training. This learning paradigm focuses on sample selection strategies, and the model training process starts from scratch. In the context of multi-source unsupervised domain adaptation (MUDA), traditional AL methods may select anomalous or redundant instances during the selection process of target domain data due to the presence of domain shift. This contradicts our intention, as we aim to enhance MUDA performance with minimal labeling costs. Some studies \citep{AADA,ADMA,CLUE} have combined active learning with domain adaptation, but these methods primarily address single-source problems and are not effective in instance selection under multi-source settings.

Based on the aforementioned issues, this paper proposes the following points in the context of multi-source domain adaptation: 1) Pursuing overall alignment between multiple source domains and the target domain is not necessarily the best strategy for this problem. To preserve the target domain's classification features as much as possible, it is essential to control the degree of feature alignment between each source domain and the target domain by weighting them based on the level of distribution differences. 2) Effective utilization of the similarity between the local distributions of different source domains and the target domain should be considered. In the process of multi-source domain adaptation, the adaptation weights for multiple domains should be dynamically adjusted based on the local distribution differences of multiple source domains to achieve an optimal learning process. 3) In the active learning sampling process, it is not sufficient to select instances based solely on the specific relationship between a single source domain and the target domain. This would result in the selection of samples biased towards a single source domain and lead to redundant sampling.

Based on the aforementioned points, this paper introduces a Dynamic Domain Discrepancy Adjustment for Active Multi-Domain Adaptation (D$^{3}$AAMDA) method that leverages the differences between the source domain and target domain distributions. Our approach fully considers the diversity of source domain data distributions during domain adaptation by weighting the alignment of features between each source domain and the target domain based on the degree of distribution differences. We establish a dynamic objective function learning convergence mechanism. Furthermore, we propose a Multi-source Active Boundary Sample Selection (MABS) strategy. This strategy implicitly expands the boundaries between different source domain classes and amplifies the differences between similar samples across classes. Based on the dynamic domain difference modulation strategy, we select target domain samples that contain the most informative content by considering the feature relationships between difficult samples from the target domain and the source domains. These selected samples are then labeled and incorporated into the model training process. In simple terms, our active learning sample selection mechanism not only focuses on the uncertainty and diversity of the target domain data itself \citep{AADA,CLUE} but also considers the relevance between difficult-to-classify samples from the source domains and the target domain data. At the same time, we ignore the redundancy brought by well-classified samples in the target domain sampling process.

The main contributions of our proposed method are as follows:
(1) Addressing the impact of complex distribution imbalance in multi-source domain data on domain adaptation learning, we construct an adaptive learning model with a dynamic domain difference modulation strategy. We define a weighted form of the multi-source deep domain adaptation objective loss function and take into account both the overall and local distribution differences between each source domain and the target domain. We propose a dynamic domain difference modulation mechanism that dynamically adjusts the weights of each source domain in the overall loss function.
(2) We introduce a Multi-source Active Boundary Sample Selection (MABS) strategy that enhances the role of difficult samples in shaping decision boundaries by employing a dynamic boundary loss. This strategy guides the model to pay more attention to the impact of difficult examples on the selection of target domain samples, by focusing on the difficulties in the classification boundaries of each source domain. Our sampling strategy aims to improve model performance with minimal annotation costs, and to the best of our knowledge, this is the first work that applies active learning to the multi-source domain adaptation problem.
(3) We evaluate the performance of our method on multiple domain adaptation benchmark datasets. The experimental results demonstrate that our approach achieves state-of-the-art performance.

\section{Related Work}
\label{Related Work}

\textbf{Domain Adaptation (DA)} aims to generalize models trained on the source domain to the target domain, considering the differences in their feature distributions. Early methods in DA focused on extracting domain-invariant features to achieve this goal. In recent years, some DA methods have addressed domain shift by adjusting the feature distributions between the source and target domains. For example, certain works \citep{a18,D-CORAL,a22,a23,a50} minimize various metrics to align the features of the source and target domains. Other works adopt adversarial approaches, where the training process involves a game between a feature extractor and a domain discriminator. The domain discriminator learns to differentiate between source and target domain features, while the feature extractor learns to confuse the domain discriminator by extracting domain-invariant features, ultimately aligning the features between the source and target domains. For instance, \cite{a24} proposed Domain-Adversarial Neural Network (DANN) based on Generative Adversarial Networks (GAN) \citep{a53}. \cite{a26} transformed the adversarial conditions into domain discrimination information based on the correlation between classes and expanded the conditional adversarial mechanism. \cite{a57} proposed an integrated classifier for both class and domain classification to align the joint feature distributions across domains. \cite{a25} categorized feature representations into shared and private domains, modeling the uniqueness of each domain to enhance the ability to extract domain-invariant features. The above-mentioned methods aim to minimize the differences in feature distributions between the source and target domains to achieve domain alignment and subsequently employ source domain classifiers for discriminating target domain data.

\textbf{Multi-Source Domain Adaptation (MSDA)} allows for different distributions in the source domains and addresses the target domain discrimination problem by combining multiple source domains. However, the core idea of MSDA still involves aligning the various domain data through a shared network backbone \citep{a14,a61,a13,a16,MFSAN,a30}. Some methods align domain distributions through weighting schemes. For example, \cite{a37} proposed distribution-weighted combination rules, where the target domain distribution is obtained as a weighted combination of multiple source domain distributions. By leveraging the relationships between the source domains and the target domain, using multiple source classifiers can yield optimal predictions for the target domain. \cite{a60} explicitly learned the matching degree between source and target domain samples using meta-training based on point-to-set metric criteria to update the weight parameters for different source domains in relation to the target domain. \cite{a45} employed non-shared feature extractors to preserve more domain-discriminative features and achieved recognition on the target domain through weighted combinations of source classifiers. Other methods, such as M$^3$SDA proposed by \cite{a14}, align the source domains with each other while aligning them with the target domain. Additionally, they collected and annotated a new domain adaptation dataset called DomainNet. \cite{a44} utilized curriculum learning and adversarial learning to select the most suitable source domain samples for alignment with the target domain distribution. \cite{a43} constructed a knowledge graph based on the prototypes of each source domain to aggregate existing knowledge from multiple domains, guiding the prediction of target domain samples. \cite{STEM} employed a joint feature extractor and combined multiple teachers in a mixture of source domains for global predictions.

\textbf{Active Domain Adaptation (ADA)} incorporates the idea of Active Learning (AL) \citep{B1,B2,B3,B4}, which aims to enhance model performance by actively selecting the most informative data for annotation, minimizing the annotation cost. Existing AL methods mostly rely on sample uncertainty \citep{B5,B6,B7} and representativeness \citep{B8} to assess the annotation value of samples. ADA methods draw inspiration from AL and aim to improve domain adaptation with limited annotations. Most ADA methods design query functions to rank the annotation value of target domain samples. For example, \cite{AADA} applying AL to domain adaptation, represents targets based on the scores of domain discriminators and constructs a sample query function. \cite{TQS} proposed a random selection strategy to enhance sample diversity in the sample selection process. Another category of methods models the annotation value of target domain data based on clustering ideas. For instance, \cite{CLUE} selects samples using uncertainty-weighted clustering within a clustering framework. These methods have achieved good results, but there are still limitations in considering certain aspects. For example, they do not take into account the domain similarity between different source domains and the target domain, as well as the inter-class similarity between different source domain classes and the target domain classes, which can lead to sample redundancy. It is worth noting that, unlike most AL and ADA methods, in our approach, the annotation of target domain data is achieved through pseudo-labeling rather than using their true labels. Fundamentally, it remains an unsupervised domain adaptation method.

\section{Methodology}
\label{Methodology}
\subsection{Problem definition}

In our approach, we consider M labeled source domain datasets $S=\{S_1,S_1,S_1,\ldots,S_M\}$ and one unlabeled target domain dataset $T$. Each source domain $S_M$ contains $N_{S_M}$ samples and label pairs ${\{(x_i^{S_M},y_i^{S_M})\}}_{i=1}^{N_{S_M}}$, while the target domain data $T$ only consists of $N_T$ samples $X_T=\{x_i^T\}_{i=1})^{N_T}$, with unknown labels. However, the class space is shared among the source domains and the target domain, i.e., $C_{S_1}=C_{S_2}=\ldots=C_{S_M}=C_{S_T}$, where $C$ represents the class index ranging from 1 to $K$, and $K$ is the number of classes.

In the multi-source active sampling process, we first initialize the labeled target domain sample set ${T_L=\emptyset}$. After training the model for a specified number of epochs, the model performs active sampling based on a sampling rate $\Delta$. Then, we assign pseudo-labels $y_t^{'}$ to the selected target domain samples $x_t^{'}$, forming labeled target domain data ${(x_t^{'},y_t^{'})}$ that is added to $T_L$, and remove this data from the target domain $T$, denoted as $T=x_i^T{/}x_t^{'}$. Finally, we train the model using the combined dataset $S\cup T_L$ and repeat the active sampling process in the next specified epoch.

Our objective is to fully consider the distinct distribution differences among multiple source domains and the target domain within the domain adaptation learning process. To achieve this, we propose a dynamic weight adjustment mechanism that adapts to these differences. Furthermore, we employ active learning techniques to mitigate negative transfer during the domain adaptation process, while minimizing the cost of annotation.

\subsection{Method framework}

In this paper, we first define a weighted domain discrepancy loss function based on the distribution differences between each source domain and the target domain. We incorporate a dynamic weight adjustment mechanism during the learning process to control the adaptive learning weights of each source domain. Furthermore, we enhance the influence of challenging samples from the source domains on the decision boundary through a dynamic boundary loss, resulting in a clearer decision boundary. Based on this, we design a target domain sample query function for the active sampling strategy to assess the importance of unlabeled target samples. The proposed approach consists of two components: the dynamic domain discrepancy adjustment algorithm and the multi-source active boundary sample selection strategy.

\textbf{Dynamic domain discrepancy adjustment algorithm} consists of loss functions for each domain and their corresponding dynamic weight adjustment modules. Based on the overall distribution distances between the source domains and the target domain, the weights of the domain losses are dynamically adjusted for local samples in each training iteration. This adjustment aims to reduce loss oscillation and allow the target domain data to align more closely with the source domain samples that have higher relevance, promoting positive transfer and further enhancing model generalization.

\begin{figure*}
	\centering 
	\includegraphics[width=1\textwidth, angle=360]{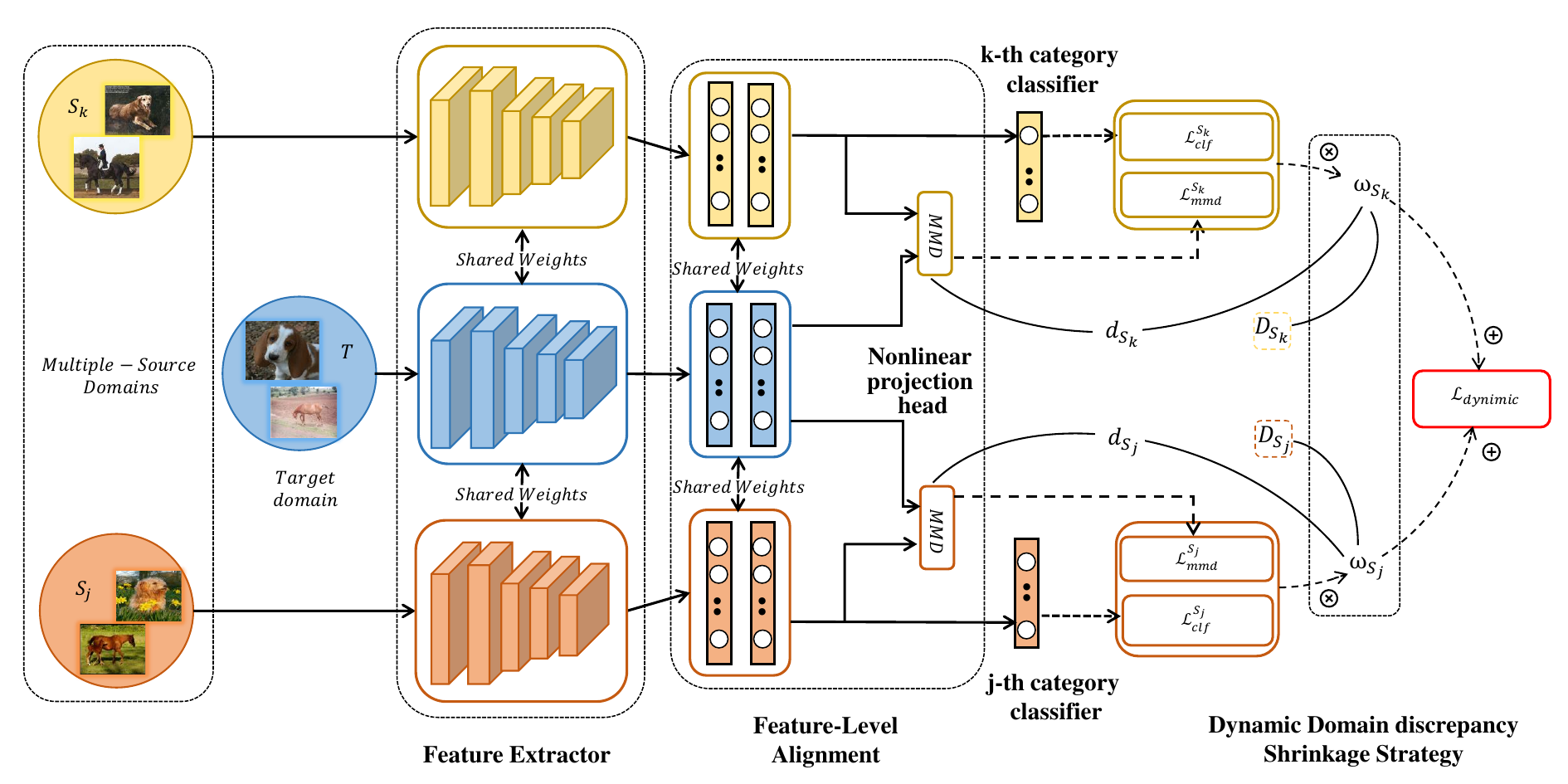}
    \setlength{\abovecaptionskip}{2pt}%
	\caption{ Overall framework of Dynamic Domain Discrepancy Adjustment algorithm. It contains four parts: shared feature extraction, nonlinear projection head, domain alignment, and multi-source classifier.} 

\end{figure*}

we employ ResNet as the shared feature extractor. To mitigate the loss of metric features caused by the classification loss function and integrate the information extracted from different feature layers, we introduce a non-linear projection head $g(\cdot)$ after the shared feature extractor. This enhances the quality of representations extracted by the preceding layers of the network, thereby avoiding the issue of feature omission when measuring the domain discrepancy in subsequent steps. The non-linear projection head in our approach consists of two fully connected layers, as shown in Equation (1), where $\sigma$ denotes the ReLU activation function. The dimensionality of the non-linear projection head is consistent with the output dimensionality of the shared feature extractor.

\begin{equation}
    z = g(F(\cdot))=W^{(2)}\sigma(W^{(1)}F)
\end{equation}

The Domain-Level Alignment component employs the Maximum Mean Discrepancy (MMD) to assess the domain differences. Equation (2) represents the MMD in the Reproducing Kernel Hilbert Space (RKHS), where $\mathcal{H}$ denotes the space associated with the kernel function. $F$ represents the mapping function that maps the original features to the RKHS.

\begin{equation}
    D_{\mathcal{H}}({F,p,q}) \triangleq sup_{f \in F}(\mathbb{E}_{x \sim p}[f(x)]-\mathbb{E}_{y \sim q}[f(y)])
\end{equation}

The domain discrepancy is quantified using Equation (3). $\widehat{D}_H^{S_i}(p_{S_i},p_t)$ represents the unbiased estimation of $D_H^{S_i}(p_{S_i},p_t)$, where $p_{S_i}$ and $p_t$ denote the feature distributions of the source domain and target domain, respectively. The kernel function $\phi(\cdot)$ employed in the calculation utilizes a Gaussian kernel. By calculating the MMD distances between the feature distributions of each source and target domain, and incorporating them into the loss function, the distributions of the source and target domains are progressively aligned through iterative optimization. The domain discrepancy loss is expressed as Equation (4).

\begin{equation}
\begin{aligned}
    \widehat{D}_{\mathcal{H}}^{S_i}(X_{S_i},X_T) &\triangleq  \\
    &\Vert {\frac1{N_{S_M}}\sum_{x^{S_i} \in X_{S_i}}[\phi(x^{S_i})]-\frac1{N_T}\sum_{x^T \in X_T}[\phi(x^T)]} \Vert_{\mathcal{H}}^2 
\end{aligned}
\end{equation}

\begin{equation}
    \mathcal{L}_{mmd}^{S_i}=\widehat{D}_{\mathcal{H}}^{S_i}((z(X_{S_i}),z(X_T))
\end{equation}

Network maps the data from multiple source domains and the target domain to a shared feature space using the shared feature extractor $F$. During the training process, the network aims to minimize the distance between the feature distributions of the source domains and the target domain, as indicated by Equation (4). The cross-entropy loss function $\mathcal{L}_{clf}^{S_i}(F,C)$, defined in Equation (5), is used as the classification loss to optimize both the classifier $C$ and the feature extractor $F$, $N_{S_i}$ represents the size of the sample set in the source domain $S_i$, $y_j^{S_i}$ denotes the true label of the j-th sample $x_j^{S_i}$ in the source domain $S_i$, and $\widehat{y}_j^{S_i}\in (0,1)$ represents the corresponding softMax prediction result.

\begin{equation}
    \mathcal{L}_{clf}^{S_i}(F,C)=-\mathbb{E}_{(X_{S_i},Y_{S_i})\sim p_{S_i}} \frac1{N_{S_i}}\sum_{j=1}^{N_{S_i}}\sum_{k=1}^K{y}_j^{S_i} \log \widehat{y}_j^{S_i}(k)
\end{equation}

\textbf{Dynamic Weight Adjustment}. In the process of domain adaptation, it is crucial to consider the distribution distances between source domains and the target domain in order to determine the weight parameters guiding the learning of the target domain. However, assigning equal weights to all samples within the same source domain and minimizing the domain discrepancy across all domains may result in negative transfer when certain source domain samples significantly differ from the target domain. To mitigate this issue, we propose a dynamic weight adjustment that adaptively adjusts the weights of source domains based on their distribution distances from the target domain. Through iterative modifications of the specific influence of each source domain, our algorithm dynamically selects source domain samples that are most relevant to the target domain samples, aiming to reduce the domain discrepancy between source and target domains and improve the accuracy of target domain sample classification.

Based on the pre-trained feature extraction network, we compute the overall distribution distances $\{D_{S_1}, D_{S_2}, ..., D_{S_M}\}$ from each source domain $\{S_1, S_2, ..., S_M\}$ to the target domain, as described in Equation (3). Assuming that the distribution distance between the source domain $k$ and the target domain is the smallest, we apply an inverse transformation to the overall distribution distances and the distance between source domain $k$ and the target domain$\{\frac{D_{S_k}}{D_{S_1}}, \frac{D_{S_k}}{D_{S_2}}, \cdots, 1, \cdots, \frac{D_{S_k}}{D_{S_M}}\}$, resulting in $\{\alpha_{S_1}, \alpha_{S_1}, \cdots, \alpha_{S_k}, \cdots, \alpha_{S_M}\}$. In each mini-batch, We first compute the batch-wise distribution distances $\{d_{S_1}, d_{S_2}, ..., d_{S_M}\}$, by minimizing the domain discrepancy loss, we aim to narrow the distribution gaps between each source domain and the target domain. Based on the overall distances, we calculate the dynamic adjustment factor $\omega_{S_i}$ for source domain $i$ using Equation (6), where $\epsilon_i$ is computed as $\frac{e^{-d_{S_i}}}{\sum_{j=1}^Me^{-d_{S_j}}}$. The sign of $\epsilon_i$ is determined by comparing the magnitudes of $\frac{D_{S_i}}{\sum_{j=1}^MD_{S_j}}$ and $\frac{d_{S_i}}{\sum_{j=1}^Md_{S_j}}$. If $\frac{D_{S_i}}{\sum_{j=1}^MD_{S_j}} > \frac{d_{S_i}}{\sum_{j=1}^Md_{S_j}}$, $\epsilon_i$ is positive; otherwise, negative. During the domain adaptation training, the batch-wise distribution distances are used to adjust the impact strength of the overall distribution distances.

\begin{equation}
    \omega_{S_i}=\alpha_{S_i}\pm\epsilon_{S_i}
\end{equation}

Taking the source domains $S_k$ and $S_j$ as an example, the dynamic weight adjustment mechanism determines the dynamic adjustment factors $\omega_{S_k}$ and $\omega_{S_j}$ for source domains $k$ and $j$, respectively, based on the overall distribution distances $D_{S_k}$ and $D_{S_j}$ between the source domains and the target domain. These factors are computed using the batch-wise adjusted distances $d_{S_k}$ and $d_{S_j}$ during the iterative process. By incorporating the dynamic adjustment factors, the network can leverage more relevant source domain sample features in each training iteration, allowing the target domain to benefit from the information provided by the source domains that are more closely related to it. This dynamic adjustment enables the target domain to effectively utilize the features of the related source domain samples during every training process.

The dynamic domain discrepancy loss function is defined as equation (7), where $\beta$ is a hyperparameter that controls the weight of domain adaptation and its optimal value for the single-source domain scenario is determined empirically. $\mathcal{L}_{clf}^{S_i}(\cdot)$ represents the classification loss for the i-th source domain, and $\mathcal{L}_{mmd}^{S_i}(\cdot)$ represents the domain discrepancy loss for that source domain. $\omega_{S_i}$ is the dynamic adjustment factor for the source domain $S_i$.

\begin{equation}
    \mathcal{L}_{dynamic}=\sum_{i=1}^M\omega_{S_i}(\mathcal{L}_{clf}^{S_i}(F,C)+\beta\mathcal{L}_{mmd}^{S_i})
\end{equation}

\textbf{Multi-source Active Boundary Sample Selection strategy}.
To alleviate the source domain bias, we introduce a simple dynamic boundary loss term, as shown in Equation (8). The term $[C(z(x))_i-C(z(x))_y+d]_{+}$ ensures that the larger value between 0 and $[C(z(x))_i-C(z(x))_y+d]$ is taken, where d is a hyperparameter that determines the desired boundary. Here, $C(\cdot)_i$ and $C(\cdot)_y$ represent the outputs of the classifier for the i-th class and the true class, respectively. By incorporating Equation (8) into the training process, the model focuses more on difficult samples that are challenging to discriminate in each source domain, thus enhancing the model's classification capability. This dynamic boundary loss term encourages the model to pay greater attention to samples near the decision boundaries, which leads to an improved decision boundary and overall classification performance.

\begin{figure*}
	\centering 
	\includegraphics[width=1\textwidth]{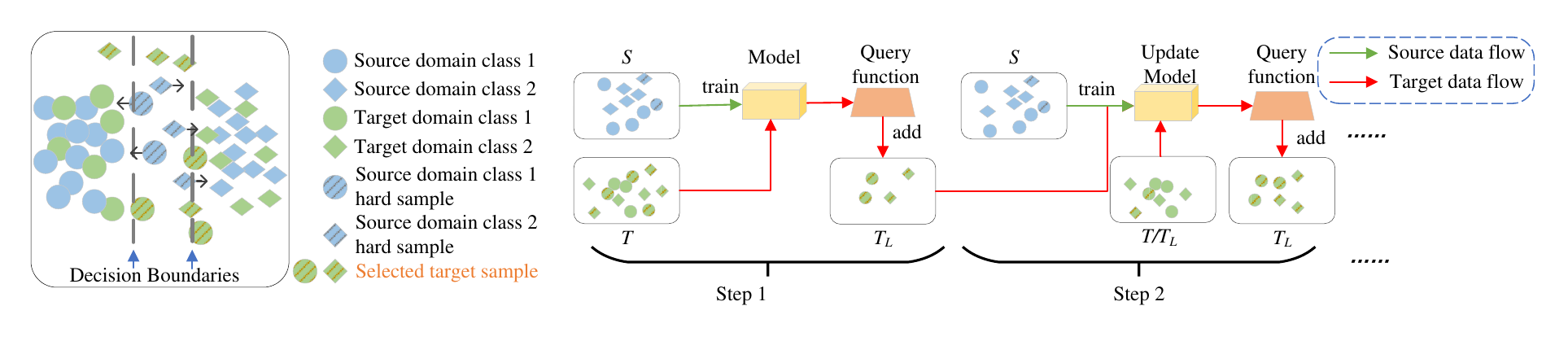}
    \setlength{\abovecaptionskip}{2pt}%
	\caption{ Active sample screening strategy. Firstly, the hard samples of each category are kept away from the classification boundary by dynamic boundary loss, which in turn expands the decision boundary between categories, and secondly, the samples in the target domain data that are at the decision boundary are screened and labeled by our query function to expand the labeled target domain sample set $T_L$.} 

\end{figure*}

\begin{equation}
    \mathcal{L}_{dis}(x^{S_i},y^{S_i})=\sum_{k=1}^K \omega_{S_i}[C(z(x^{S_i}))_k-C(z(x^{S_i}))_y+d]_{+}
\end{equation}

The dynamic boundary loss allows us to determine that for the target domain data, the smaller the discrepancy in the classification outputs among different classes obtained by the softMax layer, the greater the influence of that data on the decision boundary. Based on this characteristic, we design a query function, as shown in Equation (9), to obtain the ranking of sample importance in the target domain.

In the query function, when a sample exhibits more balanced predictions among different classes, indicating higher uncertainty in the model's predictions, the sample is assigned a higher importance score. The term $softMax(C(z(x^T )))_i$ represents the i-th value in the logit vector obtained by sorting the output of the softmax layer in descending order.

\begin{equation}
    I(x^T)=1-[softMax(C(z(x^T)))_1-softMax(C(z(x^T)))_2]
\end{equation}

We can sort the samples from the target domain based on their importance scores obtained from Equation (9). Then, we select the top-ranked samples according to a given sampling rate, assign pseudo-labels to these samples, and include them in the labeled target domain sample set, $T_L$. Finally, we update the model using the combined dataset, $T_L\cup S$, where $S$ represents the labeled source domain samples.

\section{Experiments}
\label{Experiments}

To verify the effectiveness of our method, we draw on the dataset used in the literature related to active domain adaption \citep{AADA,TQS,CLUE} to validate the performance of our method and further analyze the experimental results.

\subsection{Data description}

we conducted performance testing of our method using the Office-31, Office-Caltech 10, and Office-Home datasets.
The Office-31 dataset \citep{a10} consists of 4,652 images collected from office environments. It includes three domains: Amazon (A), DSLR (D), and Webcam (W). Each domain contains the same 31 categories. In our experiments, we selected one domain as the target domain and used the other two domains as the source domains, creating three transfer tasks.
The Office-Caltech 10 dataset \citep{a41} is derived from the Office-31 dataset and the Caltech-256 dataset. It contains a total of 2,533 images from four domains: Amazon (A), DSLR (D), Webcam (W), and Caltech (C). The dataset consists of 10 common categories across domains. We selected three subsets from the Office-Caltech 10 dataset as the source domains and used the remaining subset as the target domain, creating four transfer tasks.
The Office-Home dataset \citep{a8} consists of four domains: Art (A), Clipart (C), Product (P), and Real-World (R). It contains a total of 65 categories, and the category labels are consistent across domains. We selected three subsets from the Office-Home dataset as the source domains and used the remaining subset as the target domain, creating four transfer tasks.

\subsection{Experiment setup}

We conducted several comparative experiments in this section using different datasets. Firstly, we compared our approach with a baseline where only the source domain data is used for training. Secondly, we compared our method with several classical single-source domain adaptation methods, including Transfer Component Analysis (TCA) \citep{a29}, Geodesic Flow Kernel (GFK) \citep{a11}, Deep Domain Confusion (DDC) \citep{a18}, Deep CORAL \citep{D-CORAL}, Reversed Gradient (RevGrad) \citep{a31}, Domain Adaptation Network (DAN) \citep{a19}, Residual Transfer Network (RTN) \citep{a32}, Joint Adaptation Network (JAN) \citep{a1}, Manifold Embedded Distribution Alignment (MEDA) \citep{a2}, Maximum Classifier Discrepancy (MCD) \citep{a3}, Adversarial Discriminative Domain Adaptation (ADDA)\citep{a27}, CDT \citep{CDT}, FixBi \citep{fixbi}, and single-source active domain adaptation methods such as UCN\citep{UCN}, QBC\citep{QBC}, Clustering\citep{Clustering}, AADA \citep{AADA}, ADMA \citep{ADMA}, TQS \citep{TQS}, and CLUE \citep{CLUE}.

For the single-source domain adaptation experiments, we employed two strategies. The first strategy is the Single-Best method, where each source domain is independently adapted to the target domain, and the highest recognition accuracy among the source domains is chosen as the final result. The second strategy is the Source-Combine method, where all the source domain data is combined into a single dataset, and a single-source domain adaptation method is applied to solve the multi-source adaptation problem. This strategy aims to explore whether incorporating data from other source domains can improve the performance of single-source domain adaptation.

Finally, we compared our method with several state-of-the-art multi-source deep domain adaptation methods, including Deep Cocktail Network (DCTN) \citep{a16}, Multi-Source Distilling Domain Adaptation (MDDA) \citep{a45}, Multi-Source Domain Adversarial Networks (MDAN) \citep{a13}, Moment Matching for Multi-Source DA (M$^3$SDA) \citep{a14}, Learning to Combine for Multi-Source Domain Adaptation (Ltc-MSDA) \citep{a43}, SImpAI\citep{SImpAI}, DARN\citep{DARN}, SHOT\citep{SHOT}, MFSAN\citep{MFSAN}, STEM\citep{STEM}, DINE\citep{DINE}, PTMDA\citep{PTMDA}, and Curriculum Manager for Source Selection (CMSS) \citep{a44}.

\begin{table}[h]
\resizebox{1\columnwidth}{!}{
  \centering
  
    \begin{tabular}{llcccc}
    \toprule Method 
        & Model & A, D→W & A, W→D & D, W→A & Avg \\
    \midrule
    \multicolumn{1}{l}{\multirow{8}[1]{*}{Single-best}} 
        & Source Only & 96.7  & 99.3  & 62.5  & 86.2 \\
        & TCA   & 96.7  & 99.6  & 63.7  & 86.7 \\
        & GFK   & 95.0    & 98.2  & 65.4  & 86.2 \\
        & DDC   & 95.2  & 98.2  & 67.4  & 86.9 \\
        & CORAL & 98.0    & 99.7  & 65.3  & 87.7 \\
        & RevGrad & 96.9  & 99.2  & 68.2  & 88.1 \\
        & DAN   & 96.8  & 99.6  & 66.8  & 87.7 \\
        & RTN   & 96.8  & 99.6  & 66.2  & 87.5 \\
        & FixBi & 99.3  & 100.0   & 79.4  & 92.9 \\
        & CDT   & 99.0    & 100.0   & 76.5  & 91.8 \\
    \midrule
    \multicolumn{1}{l}{\multirow{3}[2]{*}{Source-combine}} 
        & RevGrad & 98.1  & 99.7  & 67.6  & 88.5 \\
        & CORAL & 98.0    & 99.3  & 67.1  & 88.1 \\
        & DAN   & 97.8  & 99.6  & 67.6  & 88.3 \\
    \midrule
    \multicolumn{1}{l}{\multirow{7}[2]{*}{Multi-source}} 
        & DCTN  & 96.9  & 99.6  & 54.9  & 83.8 \\
        & MDAN  & 95.4  & 99.2  & 55.2  & 83.3 \\
        & MDDA  & 97.1  & 99.2  & 56.2  & 84.2 \\
        & Ltc-MSDA  & 97.2  & 99.6  & 56.9  & 84.6 \\
        & DINE  & 98.4  & 100.0   & 76.8  & 91.5 \\
        & PTMDA & 99.6  & 100.0   & 75.4  & 91.7 \\
        & \textbf{Ours}  & \textbf{100.0} & \textbf{100.0} & \textbf{81.3} & \textbf{93.8} \\
    \bottomrule
    \end{tabular}%
}
  \label{tab:addlabel}%
  \setlength{\abovecaptionskip}{2pt}%
  \setlength{\belowcaptionskip}{4pt}%
\caption{ Classification accuracy (\%) on the Office-31 dataset}
\end{table}%

Our method is implemented based on PyTorch. We fine-tune all the layers of the pre-trained ResNet-50 model trained on ImageNet using the source domain data. We also train the non-linear projection head and the classifier. We utilize the stochastic gradient descent (SGD) optimizer with a momentum of 0.9. The learning rate is set to ten times the learning rate of the feature extractor.

In addition, we apply exponential decay to the learning rate during training iterations, as shown in Equation (10), where $lr_init$   represents the initial learning rate, which is set to 0.01, $\gamma=0.8$, epoch denotes the current number of training epochs, and $epoch_{drop}$ is set to 10. Our sampling strategy takes effect from the 20th epoch onwards, with sampling performed every 2 epochs. The sampling rate is set to $\Delta=0.01$, and we conduct 5 rounds of sampling.

\begin{equation}
    lr_{new}=lr_{init}\times \gamma^{(1+epoch){/}epoch_{drop}}
\end{equation}

\subsection{Experiment results and analysis}

The experimental results of our method on the Office-31, Office-Home, and Office-Caltech 10 datasets are shown in Tables 1, 2, and 3, respectively. In the tables, "source-only" indicates applying the model trained solely on the source domain directly to the target domain for testing. "single-best" refers to using single-source domain adaptation methods to transfer each source domain to the target domain and selecting the highest accuracy as the reported result.
Furthermore, we compared our method with several advanced single-source active domain adaptation methods on the Office-31 dataset. Since our method is designed for multi-source settings, we report the average classification accuracy of the single-source methods across all source domains to the same target domain as the model's accuracy on that target domain. The experimental results are presented in Table 4.

As shown in Table 1, our method achieves state-of-the-art recognition accuracy on the Office-31 dataset. Particularly, in the challenging task where the target domain is Amazon, our method outperforms the current state-of-the-art method by 4.5\% in terms of recognition accuracy. The overall average accuracy of our method across the three transfer tasks surpasses the second-best method by 2.5\%.

\begin{table}[h]
\resizebox{1\columnwidth}{!}{
    \begin{tabular}{llccccc}
    \toprule Method 
            & Model & C,P,R→A & A,P,R →C & A,C,R→P & A,C,P→R & Avg \\
    \midrule
    \multicolumn{1}{c}{\multirow{4}[2]{*}{ Single Best}} 
            & Source only & 65.3  & 49.6  & 79.7  & 75.4  & 67.5 \\
            & DDC   & 64.1  & 50.8  & 78.2  & 75.0    & 67 \\
            & DAN   & 68.2  & 56.5  & 80.3  & 75.9  & 70.2 \\
            & RevGrad & 67.9  & 55.9  & 80.4  & 75.8  & 70 \\
    \midrule
    \multicolumn{1}{c}{\multirow{3}[2]{*}{Source Combine}} 
            & DAN   & 68.5  & 59.4  & 79.0    & 82.5  & 72.4 \\
            & Deep CORAL & 68.1  & 58.6  & 79.5  & 82.7  & 72.2 \\
            & RevGrad & 68.4  & 59.1  & 79.5  & 82.7  & 72.4 \\
    \midrule
    \multicolumn{1}{c}{\multirow{7}[2]{*}{Multi-Source}} 
            & SImpAl50 & 70.8  & 56.3  & 80.2  & 81.5  & 72.2 \\
            & SImpAl100 & 73.4  & \textbf{62.4}  & 81.0    & 82.7  & 74.8 \\
            & DARN  & 70.0    & 68.4  & 82.7  & 83.9  & 76.3 \\
            & SHOT++ & 73.1  & 61.3  & 84.3  & 84.0    & 75.7 \\
            & MDDA  & 66.7  & 62.3  & 79.5  & 79.6  & 71 \\
            & MFSAN & 72.1  & 62.0    & 80.3  & 81.8  & 74.1 \\
            & \textbf{Ours}  & \textbf{75.4} & 62.3 & \textbf{85.1}  & \textbf{84.4} & \textbf{76.8} \\
    \bottomrule
    \end{tabular}%
    }
  \label{tab:addlabel}%
  \setlength{\abovecaptionskip}{2pt}%
  \caption{ Classification accuracy (\%) on Office-Home dataset.}
\end{table}%

In Table 2, We can see our model demonstrates superior performance in the To→W, To→D, and To→C tasks. The overall average accuracy of our method exhibits a 1.4\% improvement compared to the current state-of-the-art method.

Table 3 reveals that our method achieves the highest recognition accuracy in the To→W, To→D, and To→C transfer tasks of the Office-Caltech 10 dataset. In the To→A task, our method falls only 0.1\% short of the best-performing method. Across all four transfer tasks, our method consistently achieves the highest average recognition accuracy.

\begin{table}[h]
\resizebox{1\columnwidth}{!}{
    \begin{tabular}{lcccccc}
    \toprule Method
            & Model & A, C, D→W & A, C, W→D & A, D, W→C & C, D, W→A & Avg \\
    \midrule
    \multicolumn{1}{l}{\multirow{2}[2]{*}{Source Combine}} 
            & Source Only & 99.0    & 98.3  & 87.8  & 86.1  & 92.8 \\
            & DAN  & 99.3  & 98.2  & 89.7  & 94.8  & 95.5 \\
    \midrule
    \multicolumn{1}{l}{\multirow{10}[2]{*}{Multi-Source}} 
            & Source only & 99.1  & 98.2  & 85.4  & 88.7  & 92.9 \\
            & DAN & 99.5  & 99.1  & 89.2  & 91.6  & 94.8 \\
            & DCTN & 99.4  & 99.0    & 90.2  & 92.7  & 95.3 \\
            & JAN & 99.4  & 99.4  & 91.2  & 91.8  & 95.5 \\
            & MEDA & 99.3  & 99.2  & 91.4  & 92.9  & 95.7 \\
            & MCD & 99.5  & 99.1  & 91.5  & 92.1  & 95.6 \\
            & M$^3$SDA & 99.5  & 99.2  & 92.2  & 94.5  & 96.4 \\
            & CMSS & 99.6  & 99.3  & 93.7  & 96.6  & 97.2 \\
            & STEM & 100.0   &100.0  &94.2 &98.4 & 98.2   \\
            & \textbf{Ours}  & \textbf{100.0}  & \textbf{100.0}   & \textbf{96.5}   & 98.3  & \textbf{98.7} \\
    \bottomrule
    \end{tabular}
}
  \label{tab:addlabel}%
  \setlength{\abovecaptionskip}{2pt}%
\caption{ Classification accuracy (\%) on the Office- Caltech 10 dataset.}
\end{table}%

\begin{table}[htbp]
  \centering
    \resizebox{1\columnwidth}{!}{
    \begin{tabular}{ccccc}
    \toprule
    Method & To→A  & To→D  & To→W  & Avg \\
    \midrule
    ResNet & 64.40  & 90.45 & 85.10  & 79.98 \\
    RAN   & 75.65 & 93.35 & 91.10  & 86.70 \\
    UCN   & 78.40  & 94.90  & 93.45(D→W:100) & 88.92 \\
    QBC   & 77.60  & 94.65 & 92.95 & 88.40 \\
    Cluster & 76.80  & 93.85 & 92.15 & 87.60 \\
    AADA  & 78.45 & 94.60  & 93.40(D→W:100) & 88.82 \\
    ADMA  & 79.15 & 95.00(W→D:100) & 94.15(D→W:100) & 89.43 \\
    CLUE  & 76.10  & 95.70(W→D:100) & 93.35 & 88.38 \\
    SDM   & \textbf{81.90} & 97.40(W→D:100) & 96.75(D→W:100) & 92.02 \\
    TQS   & 80.50  & 96.40(W→D:100) & 96.10(D→W:100) & 91.00 \\
    \textbf{Ours} & \textbf{81.33} & \textbf{100.00} & \textbf{100.00} & \textbf{93.76} \\
    \bottomrule
    \end{tabular}%
    }
  \label{tab:addlabel}%
  \setlength{\abovecaptionskip}{2pt}%
  \setlength{\belowcaptionskip}{4pt}%
\caption{ Classification accuracy (\%) on the Office-31 dataset with the budget of 5\% data. Since existing active domain adaptation methods are designed for single-source settings, we compare our multi-source settings by taking the average classification accuracy of each domain in the single-source method and treating it as the classification accuracy of the model for the target domain. “RAN” represents random sampling. “(D→W:100)” indicates a classification accuracy of 100 achieved under the D→W setting."}
\end{table}%

\begin{figure*}[h]
	\centering 
	\includegraphics[width=0.8\textwidth]{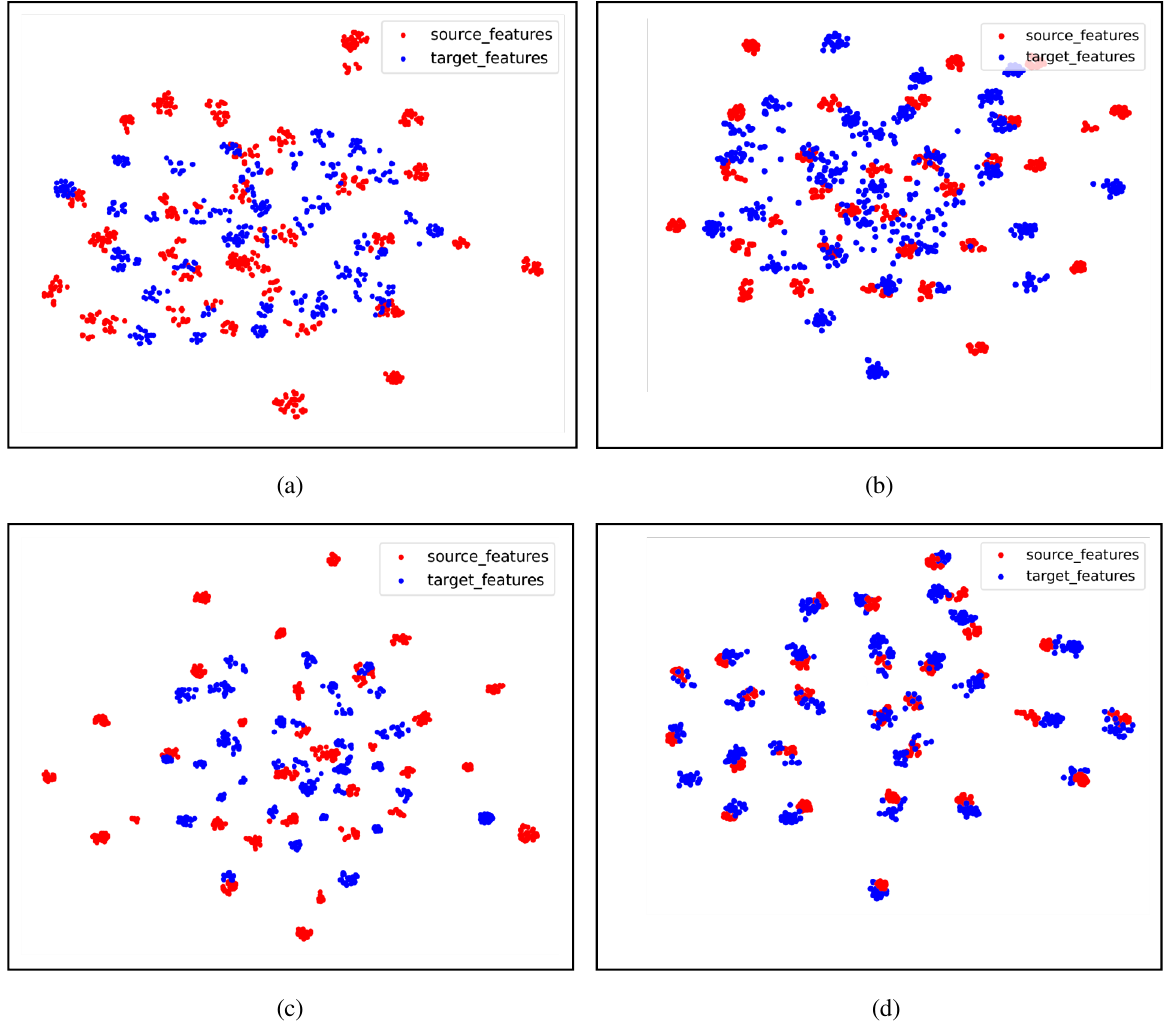}
    \setlength{\abovecaptionskip}{2pt}%
	\caption{T-SNE: Results of feature distribution visualization on the Office-31 dataset. (a), (b), (c) and (d) indicate the results of training with source domain only, adding dynamic domain difference adjustment, adding dynamic boundary loss, and the overall D$^3$AAMDA method, respectively. (d) demonstrates that the method in this paper can effectively align the source and target domain feature distributions.} 

\end{figure*}

Referring to the data in Table 4, under the constraint of the same annotation budget, our method achieves 100\% accuracy in the To→D and To→D tasks. Moreover, our method surpasses the second-best single-source method by 1.74\% in terms of overall average accuracy, thereby validating the efficacy of our proposed sampling strategy.

In addition, we visualized the clustering of latent features of the source and target domains before and after domain adaptation using T-SNE in our method. Figure 4 demonstrates that each module of our method has a positive effect on aligning the features of the source and target domains.

\subsection{Ablation analysis}

\begin{table*}[h]
  \centering
    \begin{tabular}{ccccccccc}
    \toprule
    \multicolumn{1}{c}{\multirow{3}[4]{*}{$\alpha$}} & \multicolumn{1}{c}{\multirow{3}[4]{*}{$\epsilon$}} & Nonlinear & Dynamic & Active & \multicolumn{4}{c}{\multirow{2}[2]{*}{Office-31}} \\
          &    & Projection & boundary & boundary & \\
\cline{6-7}\cline{7-8}\cline{8-9}   &    & Head & loss & sampling & A, D→W & A, W→D & D, W→A & Avg \\
    \midrule
    \checked    & ×    & ×    & ×    & ×     & 97.6  & 99.4  & 67.1  & 88.0 \\
    ×     & \checked   & \checked    & ×     & ×     & 98.5  & 99.4  & 67.5  & 88.5 \\
    \checked     & ×     & \checked     & ×     & ×     & 98.9  & 99.6  & 69.7  & 89.4 \\
    \checked     & \checked     & ×     & ×     & ×     & 98.7  & 100.0   & 68.8  & 89.2 \\
    \checked     & \checked     & \checked     & ×     & ×     & 99.0    & 100.0   & 71.6  & 90.2 \\
    \checked     & \checked     & \checked    & \checked   & ×     & 99.2  & 100.0   & 72.4  & 90.5 \\
    \checked   & \checked     & \checked     & ×     & \checked     & 99.6  & 100.0   & 75.6  & 91.7 \\
    \checked  & \checked   & \checked   & \checked   & \checked   & \textbf{100.0} & 100.0   & \textbf{81.3} & \textbf{93.8} \\
    \bottomrule
    \end{tabular}%
  \label{tab:addlabel}%
  \setlength{\abovecaptionskip}{2pt}%
  \caption{ Ablation Experiments Result. We analyze these important components by using or removing them on the ResNet-50 backbone and Office-31 datasets.}
\end{table*}%

To further investigate the effectiveness of the proposed improvements, we conducted ablation experiments. The ablation comparison modules mainly include the parameters $\alpha$ and $\epsilon$ that constitute the dynamic domain discrepancy adjustment factor $\omega$ ($\alpha$ is calculated based on the overall distribution distance between each source domain and the target domain, determining the proportion of each source domain in the overall loss function, and $\epsilon$ serves as a correction term to further adjust $\alpha$ based on the distribution distance obtained during batch training), the non-linear projection head, the dynamic boundary loss, and the proposed active boundary sample selecting strategy.
The ablation experiments led to the following conclusions: (1) The $\alpha$ determined by transforming the overall distribution distance between each source domain and the target domain can select source domains that are more similar to the target domain, which plays a major role in balancing the loss functions of different source domains. (2) The $\epsilon$ determined by transforming the batch distribution distance between each source domain and the target domain can also select source domains with higher similarity to the target domain to some extent, but it may lead to training instability and noticeable oscillations during the training process. (3) The dynamic boundary loss can further increase the distance between decision boundaries of different classes, thereby improving classification accuracy.

Furthermore, to verify the effectiveness of our active boundary sample selecting strategy, we compared it with random sampling and cluster center sampling on the Office-31 dataset. The sampling rate was set to $\Delta=0.01$, and each experiment was performed five times. The experimental results are shown in Table 6.

\begin{table}[htbp]
  \centering
  \resizebox{1\columnwidth}{!}{
    \begin{tabular}{lcccc}
    \toprule
    sampling strategy & A, D→W & A, W→D & D, W→A & Avg \\
    \midrule
    random sampling & 99.4  & 100.0   & 76.4  & 91.9 \\
    Cluster center sampling & 99.5  & 100.0   & 78.6  & 92.7 \\
    Active boundary sampling & \textbf{100.0} & 100.0   & \textbf{81.3} & \textbf{93.8} \\
    \bottomrule
    \end{tabular}%
    }
  \label{tab:addlabel}%
  \setlength{\abovecaptionskip}{2pt}%
  \setlength{\belowcaptionskip}{4pt}%
  \caption{  Impact of different sampling strategies.}
\end{table}%

The experimental results demonstrate that our active boundary sample selecting strategy can effectively select important samples that are most beneficial for target domain classification. It outperforms other sampling strategies in terms of efficiency in improving the recognition rate of the target domain. This further validates the effectiveness of our overall method in addressing the problem of multi-source unsupervised domain adaptation.

\section{Discussion}

In this paper, we propose a novel approach called Dynamic Domain Discrepancy Adjustment for Active Multi-Domain Adaptation (D$^3$AAMDA) that applies active learning to the problem of multi-source domain adaptation. D$^3$AAMDA consists of two main components. Firstly, we introduce the Dynamic Domain Discrepancy Adjustment algorithm, which dynamically adjusts the influence of each source domain on the target domain during the domain adaptation process. This allows the target domain to leverage more relevant source domain sample features. Secondly, we propose a Dynamic Boundary Loss that focuses on the contribution of difficult samples around the decision boundaries, thereby enlarging the inter-class decision boundary distances. Additionally, we incorporate an Active Boundary Sample Selecting strategy to extract important samples from the target domain, thereby enhancing the model training with minimal annotation cost. We compare our approach with several advanced domain adaptation methods, and extensive experimental analyses demonstrate the superiority of our proposed approach. In the future, we aim to extend domain adaptation methods to address the challenges of multi-modal domains and improve the generalizability of the models.






\bibliographystyle{elsarticle-harv} 
\bibliography{example}






\end{document}